\documentclass[letterpaper]{article} 
\usepackage{spconf,amsmath,graphicx}
\usepackage{times}  
\usepackage{helvet} 
\usepackage{courier}  
\usepackage{graphicx} 
\usepackage{times}
\usepackage{xcolor}
\usepackage{soul}
\usepackage{enumitem}
\usepackage[utf8]{inputenc}
\usepackage{subfigure}
\usepackage{amsmath,amssymb,amsfonts,amsthm,bbm}
\usepackage{multirow,verbatim}
\usepackage{array}
\usepackage{algorithmic}
\usepackage{algorithm}
\usepackage{xspace}
\usepackage{url}

\newcommand{\hdgc}{\emph{HyperGraphX}\xspace}

\newcommand{\gcn}{\emph{GCN}\xspace}

\newcommand{\gcnii}{\emph{GCNII}\xspace}

\newcommand{\gat}{\emph{GAT}\xspace}

\newcommand{\INPUT}{\item[\textbf{Input:}]}
\newcommand{\OUTPUT}{\item[\textbf{Output:}]}

\title{HyperGraphX: Graph Transductive Learning with Hyperdimensional Computing and Message Passing}

\name{Guojing Cong$^{\star}$ \qquad Tom Potok$^{\star}$ \qquad Hamed  Poursiami$^{\dagger}$ \qquad Maryam Parsa$^{\dagger}$ }

\address{$^{\star}$ Oak Ridge National Laboratory, Oak Ridge, TN \\
$^{\dagger}$ George Mason University, Fairfax, VA }

\begin{document}

\maketitle

\begin{abstract}
We present a novel algorithm, \hdgc, that marries graph convolution with binding and bundling operations in hyperdimensional computing for transductive graph learning. For prediction accuracy \hdgc outperforms major and popular graph neural network implementations as well as state-of-the-art hyperdimensional computing implementations for a collection of homophilic graphs and heterophilic graphs. Compared with the most accurate learning methodologies we have tested, on the same target GPU platform, \hdgc is on average 9561.0 and 144.5 times faster than \gcnii, a graph neural network implementation and HDGL, a hyperdimensional computing implementation, respectively.  As the majority of the learning operates on binary vectors, we expect outstanding energy performance of \hdgc on neuromorphic and emerging process-in-memory devices. 
\end{abstract}


\section{Introduction}

In transductive learning both labeled and unlabeled instances are present during training. As transductive approaches tailor to test instances, they often yield better performance than inductive learning, especially in data-scarce learning scenarios~\cite{joachims1999transductive}. In some applications, the labeled data constitute less than one percentage of the entire dataset. 

 Graph Neural Networks (GNNs) have become the state-of-the-art approach for learning on graph-structured data \cite{gilmer2017neural}. In a typical GNN, each node iteratively aggregates messages from its neighbors and updates its representation via learnable functions. This operation, known as message passing or graph convolution, enables nodes to encode information from their local structure and multi-hop neighborhoods, and allows the model to capture both topological and feature-based correlations. GNNs have demonstrated strong performance on tasks such as node classification and link prediction in transductive settings~\cite{kipf2016semi,velivckovic2017graph}.

Message passing in GNNs brings compute and energy challenges. Due to their inherently irregular data access patterns, GNNs incur high memory or network bandwidth requirements due to non-coalesced memory accesses on devices such as GPUs and TPUs~\cite{Wu2021Comprehensive, Zhou2020GNNSurvey} as well as frequent communication operations on clusters~\cite{Fey2021GNNEnergy, Chiang2019ClusterGCN,Zheng2020DistDGL}. These challenges motivate ongoing research into hardware-aware graph learning methods and specialized accelerators.

Hyperdimensional computing (HDC) is a brain-inspired computational paradigm that represents data as high dimensional vectors with algebraic operations such as binding and bundling \cite{Kanerva2009Hyperdimensional}. HDC vectors are noise-tolerant and support efficient computations, reflecting characteristics of neural activities in the brain with distributed representation and robustness to error. Simple operations over large binary vectors in HDC make it highly amenable to implementation on emerging energy-efficient hardware, such as in-memory computing devices and non-volatile memory~\cite{Rahimi2016Efficient}. Recently, HDC has gained some attention in graph learning~\cite{dalvi2024hdgl, gobin2024exploration, kang2022relhd,nunes2022graphhd}. Nevertheless, challenges remain with HDC transductive learning on graphs as its predictive performance oftentimes lags behind GNNs. 

We note that there are similarities in HDC graph learning and graph convolution in GNNs. For example, the vectors in both can be high-dimensional, although HDC vector dimensions tend to be much larger. Bundling in HDC graph regression is similar to read out in GNNs. In this work, we explore the cross-pollination of HDC operations with message passing of GNNs for fast, robust, and energy efficient transductive graph learning. 


Our main contributions are: 1. a new transductive graph learning algorithm that beats all GNNs and HDC approaches that we have evaluated in (average) test accuracy on our benchmarks; 2. our implementation is at least two orders of magnitude faster in training than prior approaches; and 3. we expect tremendous energy efficiency on emerging process-in-memory devices. 

\section{Preliminaries and Related work}

 Our input is a graph $G = (V, E)$, where $V$ is the set of vertices or nodes and $E$ is the set of edges. At the $l^{th}$ layer of a GNN, a feature vector $h_v^{(l)}$ is associated with each $v \in V$, and the set of hidden features for all $v \in V$ is $H^{(l)}$. $A$ is the adjacency matrix of $G$, $A=[a_{u,v}] \in \{0, 1\}$, $u,v \in V$. $D$ is the diagonal matrix with
$D_{v,v} = \sum_{u\in V} a_{u,v}$. $\hat{A}$ and $\hat{D}$ are the adjacency matrix and diagonal matrix, respectively, for graph $G$ augmented with self-loops at each node. 

\subsection{GNNs}

GNNs dominate current graph learning. They are extensively used in applications from social network analysis to materials research. For transductive learning, we evaluate milestone GNNs --  their proposals have either revolutionized learning or represented new trends in addressing prior inefficiencies.  


\noindent
\textbf{\textit{GCN: Graph Convolutional Networks.}} Kipf and Welling~\cite{kipf2016semi} propose GCN to bridge spectral graph theory with neural networks. Their key insight is to approximate spectral convolutions using localized operations, avoiding costly eigen-decomposition. GCN ignites later GNN research.  



\noindent
\textbf{\textit{GAT: Graph Attention Networks.}} To address GCN’s fixed neighborhood aggregation weights, Veličković et al.~\cite{velivckovic2017graph} propose GAT, allowing each node to assign different importance to its neighbors using attention mechanisms. GAT incorporates the defining feature of modern deep learning, that is, attention mechanism, for the first time into graph learning. 



\noindent
\textbf{\textit{GeomGCN: Geometric Graph Convolutional Network.}} While GCN and GAT rely on direct topological neighborhoods, GeomGCN \cite{Pei2020GeomGCN} is motivated by the poor performance of existing GNNs on heterophilic graphs where connected nodes often have dissimilar features or labels. GeomGCN uses a precomputed geometric neighborhood structure based on node embeddings to form a new neighborhood graph. 
GeomGCN is the first GNN to improve classification in heterophilic graphs.


\noindent
\textbf{\textit{GCNII: Deep Graph Convolutional Networks.}} GCNII \cite{chen2020simple} combats over-smoothing -- the phenomenon where node representations become indistinguishable in deep GCNs. GCNII introduces two key components: initial residual connections and identity mapping with adaptive weighting, and is one of the first GNNs to adopt more than two layers (e.g., 64 layers) of network architecture.

\subsection{HDC Graph Learning}

Kang et al. propose \textbf{RelHD}~\cite{kang2022relhd}, an HDC-based framework designed for node classification in relational and heterogeneous graphs. RelHD introduces relational encoding schemes to capture edge types and neighbor interactions.  Nunes et al.~\cite{nunes2022graphhd} introduce
\textbf{GraphHD} as a fully hyperdimensional graph classification algorithm without traditional neural architectures. It encodes each node by extracting an identifier based only on
the topology of the graph. Thus GraphHD 
uses the PageRank centrality metric. The edges are encoded by the bundling of their two endpoints.  Dalvi et al.~\cite{dalvi2024hdgl} present \textbf{HDGL}, which combines the scalability of HDC with the structural learning capabilities of message passing in GNNs. HDGL propagates hypervectors across graph neighborhoods while preserving the high-dimensional structure of the data, approximating GNN behavior in a non-neural setting.

While these studies demonstrate the potential of HDC graph learning especially for energy-efficient devices, their predictive performance is oftentimes significantly worse than sophisticated GNNs (e.g., GCNII). Also the very high dimensions of the vectors (e.g., vectors in HDGL have a dimension size of 50,000) increase memory access and communication overhead. These implementations employ floating point operations, leaving open the question whether binary operations suffice (since they can lead to even more aggressive gain on new devices). Finally, none of these implementations are evaluated on heterophilic graphs. 

\section{Incorporating message passing into HDC graph learning }
\label{s:ours}

We propose \hdgc that incorporates message passing and other modern GNN operations into HDC graph learning. \hdgc uses a high dimensional vector to represent each node. Different from prior studies, the vectors are in their original input dimension instead of much larger ones for speed and energy efficiency.

HDC employs two main operations; \emph{binding}, and \emph{bundling}. 

\noindent
\textbf{\textit{Binding ($\otimes$).}}
Binding is used to associate two hypervectors, such as a key and a value. It is typically implemented as element-wise multiplication: $\mathbf{z} = \mathbf{x} \otimes \mathbf{y}, \text{where} \quad z_i = x_i \cdot y_i$


\noindent

\noindent
\textbf{\textit{Bundling ($\oplus$).}}
Bundling aggregates multiple hypervectors into a single representative vector while preserving similarity. It is typically implemented as element-wise majority (for binary) or summation (for bipolar): $
\mathbf{z} = \mathbf{x}_1 \oplus \mathbf{x}_2 \oplus \cdots \oplus \mathbf{x}_n = \text{sign}\left( \sum_{j=1}^{n} \mathbf{x}_j \right).
$


In transductive learning for node classification, the main entities are the nodes. We bundle the local neighborhood information to a node through a weightless graph convolution: $H^{(l+1)} = \left( \hat{D}^{-1/2} \hat{A} \hat{D}^{-1/2} H^{(l)} \right)$. Compared with GCN  feature propagation $H^{(l+1)} = \sigma\left( \hat{D}^{-1/2} \hat{A} \hat{D}^{-1/2} H^{(l)} W^{(l)} \right)$, the difference is that the feature matrix $H$ in our approach does not go through the weight matrix $W$ before its aggregation. 
In HDC terms, $h^{(l+1)}_v = \sqrt{d_v}\sum_{ (u,v) \in E} h^{(l)}_u \sqrt{d_u}$. Here the bundling at a given level $l$ is done by summing the vectors of the neighboring nodes, scaled by the square root of the degree of those nodes. With $L$ layers, the feature of $v$ takes on those of nodes of the multi-hop neighborhoods, analogous to that of GNNs. In fact, we can view this as a special graph convolution without neural networks. 

To avoid over-smoothing caused by repeated aggregation resulting from large $L$, we introduce an additional bundling operation that bundles the original input vector with the vector at level $L$, as in $H = \alpha H^{(0)} + (1-\alpha) H ^{(L)}$. In HDC terms, for node $v$, we scale the two vectors, $h^{(L)}_v$ and $h^{(0)}_v$ by $\alpha$ and $(1-\alpha)$, respectively, and bundle them. Detailed description of \hdgc is shown in Algorithm~\ref{alg:hdgc}.

When node features are binary, removal of $W$  enables the use of binary vectors in the operations. When the feature vector is binary, $h_v \in \{0,1\}^{\textbf{d}}$, $\textbf{d}$ is the vector dimension, we introduce a new aggregation scheme using logical OR: $h_v^{(l+1)} = \bigvee_{(u,v) \in E} h_u^{(l)}$. This operation has several properties. First, the output feature $h_v$ remains bounded in $\{0,1\}^{\textbf{d}}$, regardless of the number of neighbors for $v$. Second, the operation is idempotent, that is, aggregating multiple times has no cumulative effect. These properties obviate the need to scale the incoming messages for each node. 
\begin{algorithm}[h]
\small
\caption{\hdgc}
\begin{algorithmic}
\INPUT Graph $G = (V, E)$, input features $\{h_v^{(0)}\}_{v \in V}$, layers $L$, parameter $\alpha$
\OUTPUT Hypervectors $\{h_v^{(L)}\}_{v \in V}$
\STATE Add reverse edges $E \gets E \cup \{(v,u) \mid (u,v) \in V\}$
\STATE Add self-loops: $E \gets E \cup \{(v,v) \mid v \in V\}$

\FOR{$\ell = 1$ to $L$}
    \FORALL{$v \in V$}
        \STATE Initialize  $m_v^{(\ell)} \gets \mathbf{0}$ 
    \ENDFOR
    \FORALL{$(u, v) \in E$}
        \IF{$h_u^{(\ell-1)}$ is binary} 
        \STATE $m_v^{(\ell)} \gets m_v^{(\ell)} \lor h_u^{(\ell-1)}$ 
        \ELSE
            \STATE $m_v^{(\ell)} \gets m_v^{(\ell)} + \frac{1}{\sqrt{d_u d_v}} h_u^{(\ell-1)}$ 
        \ENDIF
    \ENDFOR
    \FORALL{$v \in V$}
        \STATE $h_v^{(\ell)} \gets m_v^{(\ell)}$ 
    \ENDFOR
\ENDFOR
\FORALL{$v \in V$}
\STATE $h_v^{(L)} \gets \alpha h_v^{(0)} + (1-\alpha) h_v^{(L)}$
\ENDFOR
\end{algorithmic}
\label{alg:hdgc}
\end{algorithm}

At inferencing, we compute the center of the HDC vectors of training samples for each class, and for each test sample, we compare the similarity between it and each class center, and classify it to the most similar center. 
\section{Results}
\label{s:eval}

We evaluate \hdgc with both homophilic graphs and heterophilic graphs. We compare its classification (test) accuracy and training time with four GNN implementations and state-of-art HDC graph learning implementations. 
The three homophilic graphs are the citation network datasets~\cite{SNB08}: Cora, Citeseer, and Pubmed. The characteristics of these datasets are summarized in the first three columns of Table~\ref{tab:network}. These three graphs emphasize the aspect of limited training samples typical in transductive learning. The training set is very small, with only 20 labeled nodes per class. This is very different from the regular 8:1:1 or 6:2:2 train:validation:test split. For example, training samples constitute less than 1\% of the total of samples in Pubmed.  


\begin{table}[h]
\resizebox{0.48\textwidth}{!}{
\begin{tabular}{l|lll|llll} \hline
 & Cora & Citeseer & Pubmed & Chameleon & Cornell & Texas & Wisconsin \\ \hline
\# nodes & 2,708 & 3,327 & 19,717 & 2,277 & 183 & 183 & 251 \\
\# edges & 5,278 & 4,552 & 44,324 & 36,101 & 295 & 309 & 499 \\
\# features & 1,433 & 3,703 & 500 & 2,325 & 1,703 & 1,703 & 1,703 \\
\# classes & 7 & 6 & 3 & 4 & 5 & 5 & 5 \\
\hline
\end{tabular}}
\caption{Characteristics of the three homophilic citation networks, and four additional heterophilic networks}
\label{tab:network}
\end{table}


The last four graphs in Table~\ref{tab:network}, Chameleon, Cornell, Texas, and Wisconsin, are heterophilic and many GNNs struggle with these instances.  The setup here is with roughly 6:2:2 train:validation:test split, as done in prior studies with heterophilic graphs (e.g., see~\cite{CWH20}). 

All GNNs, HDGL, and \hdgc are implemented with PyTorch 2.1. \gcn, \gat, and GeomGCN implementations are taken from Deep Graph Library (DGL)
version 1.0.1~\cite{dgl}. The hyperparameters are set as those proposed by the authors in their original studies. The \gcnii implementation is by Chen \emph{et al.} with their hyperparameter setting~\cite{CWH20}.  For \hdgc there is no need to tune these hyperparameters. The only parameter to tune is $\alpha$. In our experiments we set $L=1$ for \hdgc and by default $\alpha = 0.5$. All experiments are run on NVIDIA Tesla V100S-PCIE-32GB  GPUs. 

\begin{table*}[h]
\begin{center}
\resizebox{0.8\textwidth}{!}{
 \begin{tabular}{c | c c c | c c c c| c}
 \hline
   &  Cora& Citeseer& Pubmed & Chameleon & Cornell& Texas& Wisconsin & mean \\\hline
   \gcn &  {0.815}& {0.711}& {0.790}& 0.598& 0.367& 0.408& 0.456& 0.592 \\
   \gat& 0.83& 0.725& 0.79&  0.429& 0.543& 0.583& 0.494& 0.627\\
   Geom-GCN-I& -& -& -&   0.600& 0.567& 0.575& 0.582& -\\
Geom-GCN-P&   -& -& -& 0.609&  0.608& 0.675& 0.641& -\\
Geom-GCN-S& -& -& -& 0.599&  0.556& 0.597& 0.566& -\\
   \gcnii  & \textbf{0.855} (64)& \textbf{0.734} (32)& \textbf{0.802} (16)&  0.606 (8)& 0.748 (16) & 0.694 (32) & 0.741 (16)& 0.740\\
   \hline
   HDGL& 0.752& 0.638& 0.731& 0.398& 0.575& 0.664 & 0.700& 0.636\\
   HDGL*& 0.793& 0.717& 0.767& 0.400& 0.591& 0.637& 0.694& 0.657\\
   \hdgc &  0.783 & 0.690& 0.750& \textbf{0.703}& \textbf{0.754}& \textbf{0.722}& \textbf{0.844}& \textbf{0.749} \\ \hline
 \end{tabular}}
\end{center}
\caption{Test accuracy with
  seven networks, three homophilic, with 20 training samples per each class, four heterophilic, with approximately 6:2:2 train:validation:test split. 
  Note  GeomGCN numbers in~\cite{Pei2020GeomGCN} for Cora, Citeseer, and Pubmed are with 6:2:2 splits, thus  are not included here}
\label{tab:performance-features}
\end{table*}
\begin{table*}[h]
 \begin{center}
 \resizebox{0.7\textwidth}{!}{
 \begin{tabular}{c c c c c c c c}
 \hline
   &  Cora& Citeseer& Pubmed& Chameleon & Cornell& Texas& Wisconsin \\\hline
    \gcn & 2.04 & 1.96 & 1.87&  1.87& 1.88& 1.87 & 1.88 \\ 
    \gat & 2.71 & 2.70 & 2.87&  2.64& 2.59& 2.74 & 2.63 \\ 
    Geom-GCN-I& -& -& -& 11.31& 6.53& 6.33& 6.72\\
    Geom-GCN-P& -& -& -& 9.65& 6.16& 7.04& 6.46\\
    Geom-GCN-S& -& -& -& 11.33& 6.50& 6.20& 6.33\\
     \gcnii & 104.88 & 51.67 & 12.10&  11.73& 17.62& 
     19.02& 
     15.01\\ \hline
     HDGL& 1.12& 1.17& 1.16& 1.85& 0.11& 0.12& 0.16\\
    \hdgc &  \textbf{0.0046}& \textbf{0.0130}& \textbf{0.0102}& \textbf{0.0066}& \textbf{0.0016}& \textbf{0.0013}& \textbf{0.0013}\\ \hline
 \end{tabular}}
\end{center}
\caption{Training times (seconds) for \gcn, \gat, GeomGCN, \gcnii, HDGL, and \hdgc}
\label{tab:performance-time}
\end{table*}

Transductive learning results with Cora, Citeseer, Pubmed, Chameleon, Cornell, Texas, and Wisconsin are shown in Table~\ref{tab:performance-features}. In the table, Geom-GCN-I, Geom-GCN-P, Geom-GCN-S are variants of GeomGCN with different embedding methods, Isomap (-I), Poincare (-P), and struc2vec (-S), respectively. The \gcn, \gat, and GeomGCN accuracy numbers are taken from the original publications. For \hdgc, the accuracy number for any given setting is consistent between different runs. We report the accuracy number for the heterophilic graphs as  the average of runs with 10 different splits. 

In Table~\ref{tab:performance-features} \hdgc achieves on average approximately 15.5, 12.0, and 1.0 percentage point higher accuracy than \gcn,  \gat, and \gcnii, respectively. \gcnii reports the best performance among implementations with 2, 4, 8, 16, 32, and 64 layers. The HDGL* implementation as reported in~\cite{dalvi2024hdgl} has the advantage of using both the train set and the validation set for training. Of all implementations, \gcnii has the most number of hyperparameters to tune, while \hdgc uses the most shallow networks with the least number of hyperparameters to tune. 

\hdgc exhibits outstanding performance for the four heterophilic graphs. Its accuracy is on average 29.8, 24.3, 17.4, 12.2, 17.6, and 5.8 percentage point higher than \gcn, \gat, Geom-GCN-I, Geom-GCN-P, Geom-GCN-S, and \gcnii, respectively. For Chameleon, it is about 10 percentage point more accurate than Geom-GCN-P, the next best implementation. The simplicity of \hdgc in the HDC paradigm is in sharp contrast with GeomGCN that leverages three sophisticated embedding schemes designed specifically for heterophilic graphs. \hdgc also beats \gcnii with Chameleon by about 10 percentage point, and \gcnii has a much deeper network. \hdgc achieves 0.844 accuracy for Wisconsin, again beating \gcnii with a 16 layer network by about 10 percentage point. 

We also compare the training time of \hdgc against prior implementations. The GeomGCN implementation is from~\cite{AlexFanjn_PyG_Code}.  \gcn and \gat both use a two-layer network, and are trained for 200 epochs. The \gcn hidden layer size is 16. The \gat hidden layer size is 8 with 8 attention heads. The three GeomGCN implementations use a two-layer network with a hidden layer size 64.  They are trained for a maximum of 500 epochs with a patience of 200 epochs. \gcnii uses deeper networks, and the number of layers for different graphs is the same as shown in Table~\ref{tab:performance-features}. \gcnii is trained with a patience of 100 epochs.

Table~\ref{tab:performance-time} shows the training times of \hdgc in comparison to \gcn, \gat, GeomGCN, and \gcnii. Of all GNNs, \gcn trains the fastest, followed by \gat, GeomGCN, and \gcnii. In general, the more complex (or the newer) the GNNs, the more time it takes to train. \gcnii takes more than 104 seconds to train for Cora, 50 times more than \gcn. This is mainly due to the computationally intensive back propagation with its very deep network (64 layers). In contrast, \hdgc is extremely fast. It takes a fraction of a second to compute the hypervectors. 

On average \hdgc is about 410.1 and 489.1 times faster than \gcn and \gat, respectively, for the seven graphs.  It is on average 2860.1, 2713.8, and 2811.1 times faster than Geom-GCN-I, Geom-GCN-P, Geom-GCN-S, respectively, for the four heterophilic graphs, Chameleon, Cornell, Texas, and Wisconsin. \hdgc is on average 21484.2 times faster than \gcnii for all seven graphs. \hdgc is also 144.5 times faster than HDGL.  
The minimum speedup achieved with \hdgc is 150.77 for Citeseer over \gcn, and the maximum is 22800.0 for Cora over \gcnii. 

\section{Conclusion and Future Work}

 We present \hdgc that leverages graph convolution and HDC  in transductive learning. For the seven graph benchmarks on average \hdgc beats GNN, GAT, GeomGCN, and GCNII in accuracy, and notably it performs particularly well for heterophilic graphs that are challenging for many GNNs. \hdgc achieves better accuracy than state-of-the-art HDC graph learning implementations. 

 \hdgc exhibits outstanding runtime performance in comparison to all other implementations. It is orders of magnitude faster than not only GNNs but also HDGL.  We plan to explore its implementation on emerging neuromorphic devices and also evaluate its application in graph classification tasks. 

\begingroup
\renewcommand{\baselinestretch}{0.9}\selectfont
\bibliographystyle{IEEEbib}
\bibliography{main,ml}

\begin{thebibliography}{10}

\bibitem{joachims1999transductive}
Thorsten Joachims et~al.,
\newblock ``Transductive inference for text classification using support vector machines,''
\newblock in {\em Icml}, 1999, vol.~99, pp. 200--209.

\bibitem{gilmer2017neural}
Justin Gilmer, Samuel~S Schoenholz, Patrick~F Riley, Oriol Vinyals, and George~E Dahl,
\newblock ``Neural message passing for quantum chemistry,''
\newblock in {\em International conference on machine learning}. Pmlr, 2017, pp. 1263--1272.

\bibitem{kipf2016semi}
Thomas~N Kipf and Max Welling,
\newblock ``Semi-supervised classification with graph convolutional networks,''
\newblock {\em arXiv preprint arXiv:1609.02907}, 2016.

\bibitem{velivckovic2017graph}
Petar Veličković, Guillem Cucurull, Arantxa Casanova, Adriana Romero, Pietro Lio, and Yoshua Bengio,
\newblock ``Graph attention networks,''
\newblock in {\em International Conference on Learning Representations (ICLR)}, 2018.

\bibitem{Wu2021Comprehensive}
Zonghan Wu, Shirui Pan, Fengwen Chen, Guodong Long, Chengqi Zhang, and Philip~S. Yu,
\newblock ``A comprehensive survey on graph neural networks,''
\newblock {\em IEEE Transactions on Neural Networks and Learning Systems}, vol. 32, no. 1, pp. 4--24, 2021.

\bibitem{Zhou2020GNNSurvey}
Jie Zhou, Ganqu Cui, Zhengyan Zhang, Cheng Yang, Zhiyuan Liu, Lifeng Wang, Changcheng Li, and Maosong Sun,
\newblock ``Graph neural networks: A review of methods and applications,''
\newblock {\em AI Open}, vol. 1, pp. 57--81, 2020.

\bibitem{Fey2021GNNEnergy}
Matthias Fey, Jan~E. Lenssen, Frank Weichert, and Heinrich Müller,
\newblock ``Gnnautoscale: Scalable and expressive graph neural networks via historical embeddings,''
\newblock {\em IEEE Transactions on Pattern Analysis and Machine Intelligence}, 2021.

\bibitem{Chiang2019ClusterGCN}
Wei-Lin Chiang, Xuanqing Liu, Si~Si, Yang Li, Samy Bengio, and Cho-Jui Hsieh,
\newblock ``Cluster-gcn: An efficient algorithm for training deep and large graph convolutional networks,''
\newblock in {\em Proceedings of the 25th ACM SIGKDD International Conference on Knowledge Discovery \& Data Mining}, 2019, pp. 257--266.

\bibitem{Zheng2020DistDGL}
Da~Zheng, Xiang Song, Chao Ma, Zhenyu Tan, Zhuo Ye, Jinjing Dong, Yukuo Xian, Jie Zhang, Zheng Zhang, and Alexander~J. Smola,
\newblock ``Distdgl: Distributed graph neural network training for billion-scale graphs,''
\newblock in {\em Proceedings of the 2020 USENIX Annual Technical Conference (USENIX ATC)}, 2020, pp. 33--48.

\bibitem{Kanerva2009Hyperdimensional}
Pentti Kanerva,
\newblock ``Hyperdimensional computing: An introduction to computing in distributed representation with high-dimensional random vectors,''
\newblock {\em Cognitive Computation}, vol. 1, no. 2, pp. 139--159, 2009.

\bibitem{Rahimi2016Efficient}
Abbas Rahimi, Pentti Kanerva, Luca Benini, and Jan Rabaey,
\newblock ``Efficient biosignal processing using hyperdimensional computing: Network templates for combined learning and classification of exg signals,''
\newblock in {\em Proceedings of the IEEE International Conference on Rebooting Computing (ICRC)}, 2016, pp. 1--8.

\bibitem{dalvi2024hdgl}
Abhishek Dalvi and Vasant Honavar,
\newblock ``Hyperdimensional computing for node classification and link prediction,''
\newblock {\em arXiv preprint arXiv:2402.17073}, 2024.

\bibitem{gobin2024exploration}
Derek Gobin, Shay Snyder, Guojing Cong, Shruti~R Kulkarni, Catherine Schuman, and Maryam Parsa,
\newblock ``Exploration of novel neuromorphic methodologies for materials applications,''
\newblock in {\em 2024 International Conference on Neuromorphic Systems (ICONS)}. IEEE, 2024, pp. 282--286.

\bibitem{kang2022relhd}
Jaeyoung Kang, Minxuan Zhou, Abhinav Bhansali, Weihong Xu, Anthony Thomas, and Tajana Rosing,
\newblock ``Relhd: A graph-based learning on fefet with hyperdimensional computing,''
\newblock in {\em 2022 IEEE 40th International Conference on Computer Design (ICCD)}, 2022, pp. 553--560.

\bibitem{nunes2022graphhd}
Gabriel Nunes, Abbas Rahimi, and Jan~M. Rabaey,
\newblock ``Graphhd: A hyperdimensional computing approach for graph representation and classification,''
\newblock in {\em Proceedings of the 2022 International Joint Conference on Neural Networks (IJCNN)}. IEEE, 2022, pp. 1--8.

\bibitem{Pei2020GeomGCN}
Hongwei Pei, Bingzhe Wei, Kevin Chen-Chuan Chang, Yujun Lei, and Bo~Yang,
\newblock ``Geom-gcn: Geometric graph convolutional networks,''
\newblock in {\em International Conference on Learning Representations (ICLR)}, 2020.

\bibitem{chen2020simple}
Ming Chen, Zhewei Wei, Zengfeng Huang, Bolin Ding, and Yaliang Li,
\newblock ``Simple and deep graph convolutional networks,''
\newblock in {\em International Conference on Machine Learning (ICML)}. PMLR, 2020, pp. 1725--1735.

\bibitem{SNB08}
P.~Sen, G.~Namata, M.~Bilgic, and et~al.,
\newblock ``Collective classification in network data,''
\newblock {\em AI Magazine}, vol. 29, no. 3, pp. 93--106, 2008.

\bibitem{CWH20}
M.~Chen, Z.~Wei, Z.~Huang, and et~al.,
\newblock ``Simple and deep graph convolutional networks,''
\newblock in {\em Proceedings of the 37th International Conference on Machine Learning}, Hal~Daumé III and Aarti Singh, Eds. 13--18 Jul 2020, vol. 119 of {\em Proceedings of Machine Learning Research}, pp. 1725--1735, PMLR.

\bibitem{dgl}
Y.~Rong, T.~Xu, J.~Huang, and et~al.,
\newblock ``Deep graph learning: Foundations, advances and applications,''
\newblock in {\em Proceedings of the 26th ACM SIGKDD International Conference on Knowledge Discovery and Data Mining}, New York, NY, USA, 2020, KDD '20, p. 3555–3556, Association for Computing Machinery.

\bibitem{AlexFanjn_PyG_Code}
AlexFanjn,
\newblock ``{GeomGCN\_PyG}: Geometric graph convolutional network implementation in pytorch geometric,'' \url{https://github.com/alexfanjn/GeomGCN_PyG}, 2020,
\newblock accessed: 2025-08-11.

\end{thebibliography}
\endgroup

\end{document}